\documentclass[10pt,twocolumn,letterpaper]{article}

\usepackage{ijcb}
\usepackage{times}
\usepackage{epsfig}
\usepackage{graphicx}
\usepackage{amsmath}
\usepackage{multirow}
\usepackage{url}

\usepackage[norule,symbol,perpage]{footmisc}

\usepackage{amssymb}

\usepackage[shortcuts,acronym]{glossaries}
\makeglossaries
\newacronym{pad}{PAD}{Presentation Attack Detection}
\newacronym{lbp}{LBP}{Local Binary Pattern}
\newacronym{bsif}{BSIF}{Binarized Statistical Image Features}
\newacronym{nir}{NIR}{Near-Infrared}
\newacronym{cnn}{CNN}{Convolutional Neural Network}
\newacronym{sota}{SoTA}{state-of-the-art}
\newacronym{clahe}{CLAHE}{Contrast Limited Adaptive Histogram Equalization}
\newacronym{muipad}{MUIPAD}{Mobile Uncontrolled Iris Presentation Attack Database}
\newacronym{apcer}{APCER}{Attack Presentation Classification Error Rate}
\newacronym{bpcer}{BPCER}{Bonafide Presentation Classification Error Rate}
\newacronym{ccr}{CCR}{Correct Classification Rate}
\newacronym{hter}{HTER}{Half Total Error Rate}
\newacronym{psf}{PSF}{photometric stereo features}
\newacronym{msa}{MSA}{Micro Stripes Analyses}

\ijcbfinalcopy % *** Uncomment this line for the final submission

\ifijcbfinal\fi

\makeatletter
\def\ps@IEEEtitlepagestyle{
\def\@oddfoot{\mycopyrightnotice}
\def\@evenfoot{}
}
\def\mycopyrightnotice{
{\hfill \footnotesize 978-1-7281-9186-7/20/\$31.00 \copyright 2020 IEEE\hfill}
}
\makeatother

\begin{document}

%%%%%%%%% TITLE
\title{Micro Stripes Analyses for Iris Presentation Attack Detection}

\author{Meiling Fang$^{1,2}$, Naser Damer$^{1,2}$, Florian Kirchbuchner$^{1,2}$, Arjan Kuijper$^{1,2}$\\
$^{1}$Fraunhofer Institute for Computer Graphics Research IGD,
Darmstadt, Germany\\
$^{2}$Mathematical and Applied Visual Computing, TU Darmstadt,
Darmstadt, Germany\\
Email: {meiling.fang@igd.fraunhofer.de}
}

\maketitle
\thispagestyle{empty}

%%%%%%%%% ABSTRACT
\begin{abstract}
   Iris recognition systems are vulnerable to the presentation attacks, such as textured contact lenses or printed images. In this paper, we propose a lightweight framework to detect iris presentation attacks by extracting multiple micro-stripes of expanded normalized iris textures. 
   In this procedure, a standard iris segmentation is modified. For our \gls{pad} network to better model the classification problem, the segmented area is processed to provide lower dimensional input segments and a higher number of learning samples. Our proposed \gls{msa} solution samples the segmented areas as individual stripes. Then, the majority vote makes the final classification decision of those micro-stripes. Experiments are demonstrated on five databases, where two databases (IIITD-WVU and Notre Dame) are from the LivDet-2017 Iris competition. An in-depth experimental evaluation of this framework reveals a superior performance compared with \gls{sota} algorithms. Moreover, our solution minimizes the confusion between textured (attack) and soft (bona fide) contact lens presentations.
\end{abstract}

%\let\thefootnote\relax\footnotetext{\mycopyrightnotice}
%%%%%%%%% BODY TEXT
%------------------------------------------------------------------------
\section{Introduction}
In recent years, iris recognition systems are being deployed in many law enforcement or civil applications \cite{Daugman2009,DBLP:journals/ivc/Boutros20,DBLP:conf/ijcb/Boutros20}. Compared to face authentication, iris patterns are less affected by aging \cite{Newman09}. However, recognition systems are vulnerable to presentation attacks \cite{survey18, opensource18, livedet17}. A presentation attack is performed to obfuscate the identity of the attacker or impersonate a specific person, for instance using a printed iris image, a replayed video or wearing textured contact lenses. Despite the high interest from researchers and system vendors to this issue, current algorithms still have some limitations. The LivDet-2017 Iris competition pointed out that there are still advancements to be made in the detection of iris presentation attacks, especially when unknown materials or sensors are used to generate the attacks.
Moreover, the number of smartphone users worldwide today surpasses three billion and is forecast to further grow by several hundred million in the next few years \cite{mobile_data}. With concerning the wide usage of mobile devices, there is a fact that algorithms with high computational requirements are hard to deploy in mobile devices even if it has high accuracy. 

Another challenge in iris \gls{pad} is the detection of texture contact lenses, especially when they are confused with soft (transparent) lenses. Wearing a cosmetic lens is an easy way to conceal the original texture pattern of the iris and significantly decreases the rate of genuine match rate \cite{degradation10,ndcld15, ContlensNet, livedet17}. That makes the development of a robust and accurate iris PAD algorithm becomes an essential and valuable task in real-world scenarios. Some works did specifically target the properties of the lenses by investigating spectral reflection, which summarized in \cite{survey18}. Besides, Hoffman \cite{DBLP:conf/icb/HoffmanSR19} noticed that the possible artifacts on the image dynamics around the iris/sclera border, which also occur on the contact lens presentations. They extracted eight patches from the iris-sclera boundary to capture the PA artifacts.

With these motivations, our paper targets the issue of the iris \gls{pad} and a lightweight model, and demonstrates experiments on multiple databases including two databases in the LivDet-2017 Iris competition. Our main contributions are: 1) proposed a lightweight iris \gls{pad} framework, micro-stripes analyses (\gls{msa}), that efficiently and robustly learn the image dynamics around the iris/sclera border. This system is based on a specifically trained lightweight network, and thus is suitable for deployment on mobile and embedded devices, 2) showcased that our method outperforms the \gls{sota} algorithms by comparing to all reported results on the multiple databases, 3) demonstrated the high generalization of our approach by proving its low confusion between the attack texture lenses and the bona fide soft ones.

%-------------------------------------------------------------------------
\section{Related Work}
Since deep learning methods have demonstrated superior performances in many fields like computer vision, recent iris PAD works \cite{DBLP:conf/eusipco/fang20, crossdomain19, fusionvgg18, patchcnn18} utilized neural networks to replace or fuse several traditional hand-crafted features such as \gls{lbp} \cite{lbp15, lbp14} and \gls{bsif} \cite{ndcld15, opensource18, bsif15}. Kohli \etal \cite{desist16} proposed a framework based on Multi-Order Zernike Moment and LBP features to detect presentation attacks. The fused features were classified by a three-layer neural network. Moreover, Nguyen \etal \cite{DBLP:journals/sensors/NguyenPLP18} trained specific neural networks to extract features from iris images.

Inspired by object detection methods, Chen \etal \cite{multitaskcnn18} adopted the full ocular image as input to regress parameters of the iris bounding box and compute the probability of presentation attack through a \gls{cnn} learning framework. In contrast to learning global features, Hoffman \etal \cite{patchcnn18} tried to use local features to retain as much information in the data as possible. They utilized the pre-normalized iris rather than unwrapping iris to avoid spatial information loss. However, extracting patches from the iris images lose the integrity of the iris/sclera boundary information and not all patches contain useful information for \gls{cnn}. Therefore, Hoffman \cite{DBLP:conf/icb/HoffmanSR19} selected 20 specific patches from the iris images, including eight patches from the iris-sclera boundary, to retain the useful information. Rather than only utilizing neural network features \cite{multitaskcnn18, patchcnn18}, Yadav \etal \cite{fusionvgg18} employed the fusion of Multi-level Haralick texture features with VGG \cite{vgg16} features to encode the textural variations between real and attacked iris images. Their experiment result achieved $98.99\%$ \gls{ccr} on their combined Iris database. Fang \etal \cite{fusion/Fang20} proposed a multi-layer features fusion approach based on pre-trained VGG-16 and trained from scratch MobileNet network. However, VGG-16 \cite{vgg16} contains $138$ million parameters which means it is hard to deploy on mobile devices, e.g.\ smartphones. As a follow up work on \cite{fusionvgg18}, a DenseNet based \gls{cnn} framework was proposed in \cite{densepad19} to detect presentation attacks and demonstrated its efficacy on two databases, which are captures by a mobile iris sensor in varying environmental situations. The experiment showcased good performance in detecting textured contact lens attacks. However, this method processed normalized iris images and neglected information that cross the iris boundary as will be demonstrated in this work. Moreover, Czajka \etal \cite{psf18} proposed a PAD algorithm to estimate 3D features of a pair of iris images captured from two different directions. In addition, they used the iris images with soft lenses to assess the reliability of the method. Although their method achieved good PAD performance, the experiments on one specific database are not sufficient. But, it is still valuable to assess the impact of the soft contact lens on PAD performance.
%We explore such impacts as described in Sec. \ref{ssec:experiment_setup} and results are shown in Tab. \ref{Tab:soft_lense}.

In the LivDet-2017 Iris competition \cite{livedet17}, four databases were provided for unknown type validation, and one database IIITD-WVU \cite{livedet17} was designed for cross-database validation. Chinese Academy of Sciences (CASIA) \etal \cite{livedet17} submitted a cascade SpoofNets algorithm for iris PAD in the competition and Kimura \etal \cite{spoofnet_tuning} proposed a hyperparameter tuning of the SpoofNets to improve the efficiency. However, the improvement is smaller compared to \gls{sota} algorithms. For example, Kuehlkamp \etal \cite{crossdomain19} explored combinations of CNNs with the inputs of BSIFs. They obtained superior results than the winner of the LivDet-2017 Iris competition. However, training $61$ CNNs needs high computational resources and can be seen as an over-tailored solution. With the popularity of personal smart devices, the most challenging iris \gls{pad} task is to detect presentation attacks with lower computation robustly. 

Based on the reported works, it can be noted that there is a small error gap to be closed with publicly available databases, especially with limited computational resources. In this work, we address this issue by proposing a light-weight solution that focuses on utilizes the iris/sclera boundary information.
%-------------------------------------------------------------------------
\section{Proposed Method}

\begin{figure}[t]
	\begin{center}
		\includegraphics[width=1.0\linewidth]{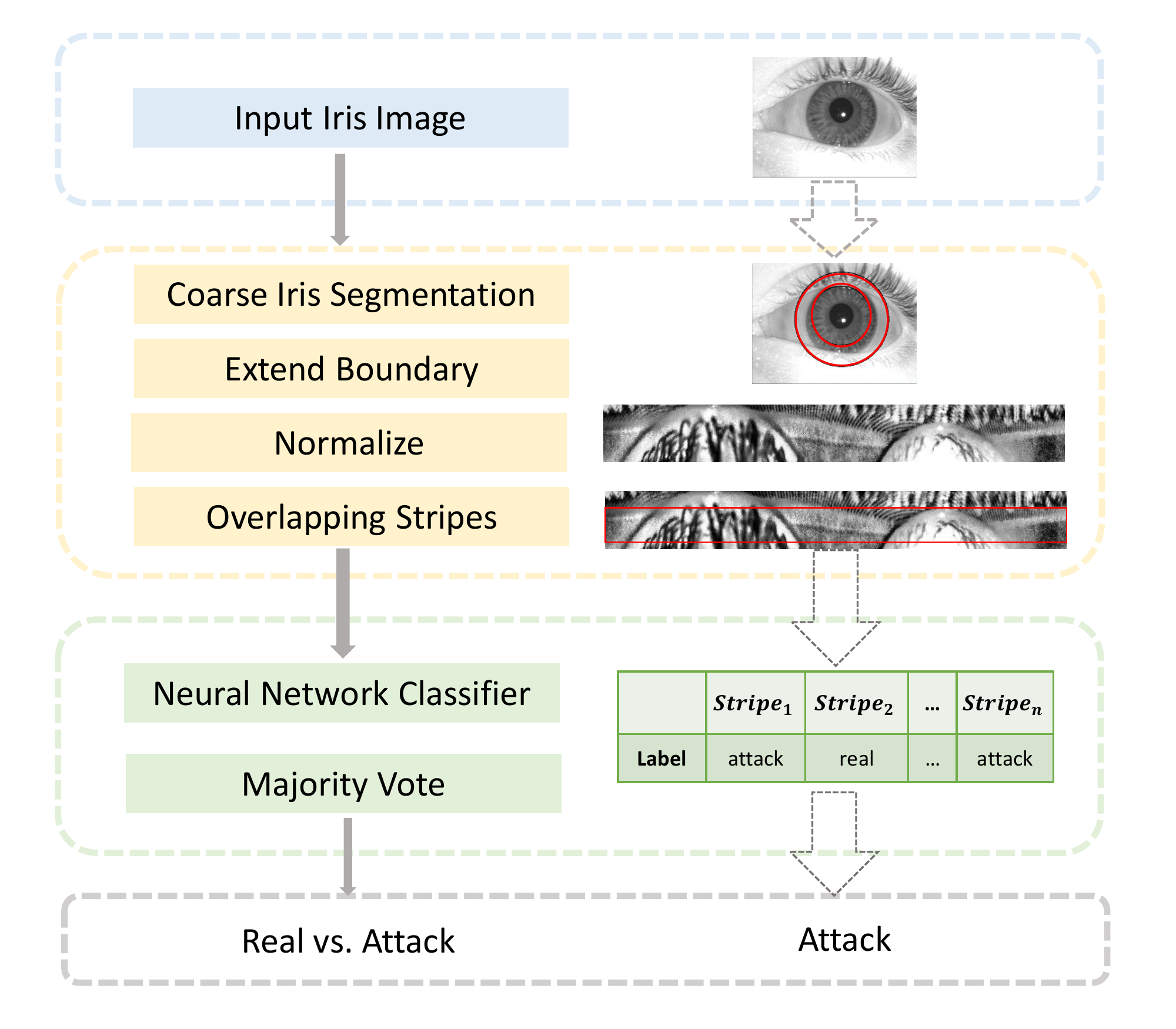}
	\end{center}
	\caption{Architecture of the proposed \gls{msa} algorithm for iris \gls{pad}. Input iris image is from NDCLD-2015 database \cite{ndcld15}.}
	\label{fig:workflow}
	%\label{fig:onecol}
\end{figure}

In this section, we describe our micro-stripes analyses \gls{msa} framework by focusing on how overlapping micro-stripes can be used to address the problem of iris presentation attack detection. Figure \ref{fig:workflow} presents the overall framework of our \gls{msa} solution. It starts with a captured iris image, the number of preprocessing steps are performed. These steps include coarse iris segmentation, segmentation extension, iris normalization and overlapping micro-stripes extraction. The processed micro-stripes are then passed into a specifically trained neural network, which results are fused in a majority vote process. The following subsections present these steps in more details. 

\begin{figure*}[t]
	\begin{center}
		\includegraphics[width=0.9\linewidth]{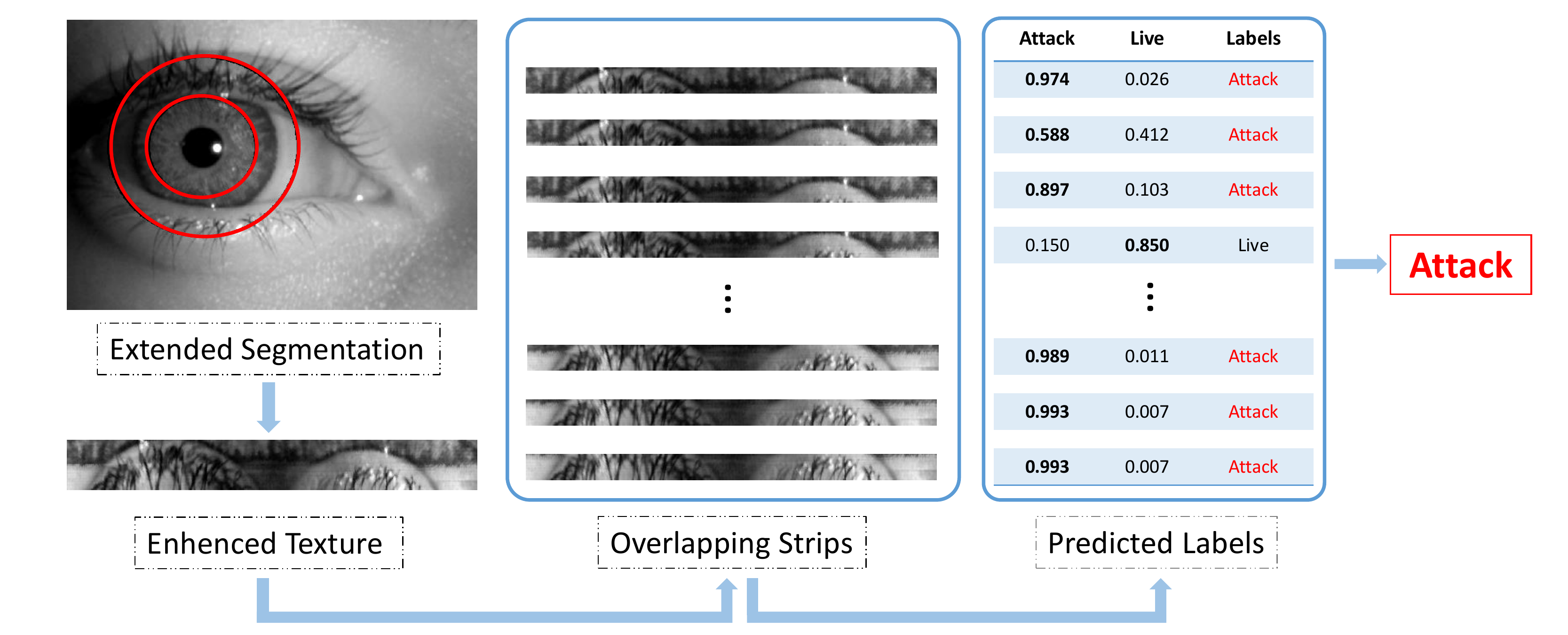}
	\end{center}
	\caption{An example illustrates the process of our \gls{msa} approach in details. in detail. The iris image with a textured contact lens comes from the IIITD-WVU database \cite{livedet17}.}
	\label{fig:stripes_samples}
\end{figure*}

\subsection{Iris Segmentation and Normalization}
Iris images do not only contain the iris region but also exhibit additional information around the iris, e.g. contact lenses edges. Previous works noticed this property and offered a solution that aims at detecting contact lens reflective properties \cite{clspectral18, leePK06}. Normally, the contact lenses cover the entire iris area and, in many cases, extend beyond it. %Given the lens attack images in public databases, one can notice the existence of a lens, see examples in Figure \ref{fig:ndcld_samples}.

First, we perform the iris coarse segmentation using the USIT v2.4.2 tool \cite{usit} to calculate the pupil and iris boundaries. In order to focus on the iris/sclera boundary region., we use the following equation to extend the pupil and iris boundaries:

\begin{equation}
	\begin{gathered}
		r_{inside} = r_{iris} - ((r_{iris} - r_{pupil}) * s_{1}) \\
		r_{outside} = r_{iris} + ((r_{iris} - r_{pupil}) * s_{2})
	\end{gathered}
\end{equation}
where $r_{iris}$ and $r_{pupil}$ represent radius of the iris and pupil. $s_{1}$ and $s_{2}$ determine the extend ratio. In experiment, $s_{1}$ and $s_{2}$ are both $\frac{2}{5}$. This aims at having an adaptive segmentation to irises of different sizes. Then, the extended segmentation is normalized using Daugman's rubber sheet expansion technique \cite{daugman04}. The normalized image is of dimension $512 \times 64$ pixels in experiment. Finally, the \gls{clahe} algorithm is applied in the normalized image to improve the contrast and enhance the texture. The segmentation and normalization processes are demonstrated in Figure \ref{fig:stripes_samples}.

\subsection{Overlapping micro-Stripes}
Once the normalized image is obtained, we extract overlapping micro-stripes for two reasons. In consonance with our assumption, the salient feature, image dynamics around iris/sclera border, in a particular stripe should indicative the belonging to bonafide or attack class. So we utilize smaller regions of texture (micro-stripes) to force the classifier to concentrating on iris/sclera area.  On the other hand, the number of training samples available in standard iris databases is limited. In our initial experiment on the full segmentation area (without stripes), we observed that the validation accuracy and loss fluctuate wildly, which indicates that the classifier suffers overfitting as demonstrated by the inferior results later. Therefore, Overlapping micro-stripes leads to lower dimensional input data and higher number of training samples, and thus better model fitting. Some of these stripes might not contain the information pattern that we are looking for as the iris/sclera and contact lens boundaries occur at different locations (in relation to the segmentation). Moreover, the iris segmentation might not always be precise. Based on these two factors, a fixed stripe is not always optimal for \gls{pad}. Therefore, we utilize multiple overlapping stripes. Each of these stripes will produce a \gls{pad} decision and the final decision will be the majority vote of these decision. The superiority of this process will be demonstrated and discussed later in the results section. 

\subsection{Classification and Fusion}
Deep learning based algorithms has been successfully used for various tasks such as classification or object detection. Recent works in the iris PAD field have already attempted to employ neural network architectures such as VGG-16 \cite{fusionvgg18}, AlexNet \cite{muipad} or custom CNN \cite{multitaskcnn18, ContlensNet} based approaches and obtained a good performance. However, despite the good detection performance, such solutions demand high computational power due to their network size, which is not acceptable for some operation scenarios. We look up a possibility to deploy the neural network with constrained hardware requirements and retain high accuracy. In contrast to VGG-16 with $138$M and AlexNet with $60$M parameters, \textbf{MobileNet V3-small} \cite{mobilenet} only has $2.5$M parameters. On that account, MobileNet V3-small is chosen as a basic structure for our proposed \gls{msa} approach to classify an iris image as a bonafide or attack. We train our own MobileNet V3 with input stripe size and use early stopping to reduce overfitting. The training-from-scratch is enabled by our overlapping micro stripes preprocessing, which provides larger training data and lower dimensionality.

Another strategy we exploit for decision-making is majority voting to enhance the robustness of our approach. The extraction of overlapping micro-stripes is not only applied for training classifier, but also used for making a final decision. Multiple micro-stripes from each texture image are naturally expected to have the same prediction by the neural network. Therefore, we sample an odd number of stripes from a normalized iris image randomly instead of using resized iris texture as input in the evaluation stage. In the end, the majority vote is employed for the final prediction.

%-------------------------------------------------------------------------
\section{Experimental Results}

\subsection{Baselines}
Three following baseline methods are implemented to analyze the results from multiple aspects: 1) hand-crafted features, 2) transfer learning of a general-purpose pre-trained network, 3) a trained from scratch computationally-efficient network.

\textbf{LBP + SVM:} LBP features, in conjunction with a Support Vector Machine (SVM) \cite{cortes1995support} using a linear kernel, are utilized to detect the presentation attacks. This method was previously proposed for iris PAD in \cite{lbp14, lbp15}. The threshold, which determines the iris image label, is chosen based on the development subset. Finally, the trained SVM model with a specific threshold is used to decide whether the iris image is bona fide or attack on the testing subset.

\textbf{VGG16 + PCA + SVM:} VGG-16 \cite{vgg16} is applied to extract iris features as used in \cite{fusionvgg18, DBLP:journals/corr/MinaeeAW17}. After extraction of VGG-16 features, a linear dimensionality reduction technique, Principal Component Analyses (PCA) \cite{doi:10.1080/pca}, is used to project the extracted feature with $7 \times 7 \times 512$ dimension into a $128$ dimensional sub-space. Those lower-dimensional features are then fed to SVM \cite{cortes1995support} and make a prediction based on the threshold choosing from the development set.

\textbf{MobileNetV3-Small:} Since we use MobileNetV3 as our basics network architecture in MSA solution, We also train it from scratch by feeding the contrast-enhanced full iris images. Again, the threshold is computed from the development set and used to predict the iris image label. The training hyperparameters are the same as ours in Tab. \ref{Tab:network_parameter}.

\subsection{Database}
\begin{table*}[!ht]
	\begin{center}
	\scalebox{0.8}{
        \begin{tabular}{|l|l|l|l|l|}
\hline
\multicolumn{2}{|l|}{Database}    & No. of training images & No. of test images & Type of Iris Images   \\ \hline
\multicolumn{2}{|l|}{NDCLD-2015 \cite{ndcld15}}  & 6,000 & 1,300  & Real, soft and textured lenses  \\ \hline
\multirow{2}{*}{NDCLD-2013 \cite{ndcld2013}}  & LG4000 & 3,000 & 1,200  & Real, soft and textured lenses   \\ \cline{2-5} 
                                  & AD100  & 600 & 300 & Real, soft and textured lenses   \\ \hline
\multirow{2}{*}{IIIT-D CLI \cite{iiitd_cli_2, iiitd_cli}} & Cognet & 1,723 & 1,785 & Real, soft and textured lenses   \\ \cline{2-5} 
                                  & Vista  & 1,523 & 1,553 & Real, soft and textured lenses    \\ \hline
\multirow{2}{*}{LivDet-2017 Competition \cite{livedet17}} & IIITD-WVU & 6,250 & 4,209 & Real, textured lenses, printouts, lens printouts \\ \cline{2-5} 
                                    & Notre Dame & 1,200 & 3,600 &  Real, textured lenses  \\ \hline
        \end{tabular}
        }
    \end{center}
	\caption{Characteristics of used databases. All databases have the training and test sets based on their own experimental setting in related papers. The Clarkson and Warsaw databases in LivDet-2017 competition are no longer publicly available.}
	\label{Tab:db_description}
\end{table*}

The proposed method is evaluated on multiple databases: three databases comprising of textured contact lens attacks captured by different sensors \cite{ndcld15, ndcld2013, iiitd_cli}, and two databases (IIITD-WVU and Notre Dame) from LivDet-2017 Iris competition \cite{livedet17}. The other two databases in the LivDet-2017 Iris competition, Clarkson and Warsaw, are no longer publicly available due to General Data Protection Regulation (GDPR) issues. The summarized information of used databases is listed in Tab. \ref{Tab:db_description}. 

\textbf{NDCLD-2013:} The NDCLD-2013 database consists of $5100$ images and is conceptually divided into two sets: 1) LG4000 including $4200$ images captured by IrisAccess LG4000 camera, 2) AD100 comprising of $900$ images captured by risGuard AD100 camera. Both the training and the test set are divided equally into no lenses (bona fide), soft lenses (bona fide), and textured lenses (attack) classes.

\textbf{NDCLD-2015:} The $7300$ images in the NDCLD-2015 \cite{ndcld15} were captured by two sensors, IrisGuard AD100 and IrisAccess LG4000 under \gls{nir} illumination and controlled environments. The NDCLD-2015 contains iris images wearning no lenses, soft lenses, textured lenses. 

\textbf{IIIT-D CLI:} IIIT-D CLI database contains $6570$ iris images of $101$ subjects with left and right eyes. For each individual, three types of images were captured: 1) no lens, 2) soft lens, and 3) textured lens. Iris images are divided into two sets based on captured sensors: 1) Cogent dual iris sensor and 2) VistaFA2E single iris sensor.

\textbf{LivDet-2017 Iris Competition Database:} LivDet-2017 Iris competition provided four databases. Since Clarkson and Warsaw are no longer publicly available, we use Notre Dame and IIITD-WVU for evaluation. The Notre Dame dataset contains images without contact lenses and with textured lenses. The IIITD-WVU dataset includes images of live irises, textured lenses, iris printouts, and printouts of textured lenses. Moreover, the Notre Dame dataset has two test sets: known attacks and unknown attacks where the contact lens manufacturers are not represented in the training set (different pattern). Notably, experiments on the IIITD-WVU dataset can be considered as cross-database evaluation where the sensors and the acquisition environments for the training and test sets are different.

\subsection{Experimental Setup} \label{ssec:experiment_setup}
\begin{table}[!ht]
\begin{center}
	\scalebox{0.9}{
        \begin{tabular}{|c|c|}
        \hline
        Parameter        & Value   \\ \hline
        No. epochs (max) & 25      \\ \hline
        Earlystopping    & 5       \\ \hline
        Learning rate    & 0.001   \\ \hline
        Optimizer        & RMSprop \\ \hline
        Batch Size        & 16 \\ \hline
        \end{tabular}
            }
    \end{center}
	\caption{Training hyperparameters.}
	\label{Tab:network_parameter}
\end{table}

Each database has predefined subject-disjoint training and testing sets. We follow this predefined division in our experiments, train on the training set, and test on the testing set. The amount of images in each set is listed in Tab 1. To provide a fair comparison, and given that the training of a neural network includes random processes leading to conversion into different optima, we repeat each of our experiments five times on the same data split and report average results f the five runs. The hyperparameters of our training model are listed in Tab. \ref{Tab:network_parameter}. We use the \textit{early stopping} with loss patience of $5$ epochs and maximum of epochs of $25$ to reduce overfitting. The coarse segmentation on every iris image is performed by USIT v2.4.2 tool \cite{usit}.

The following three experimental compositions of training/testing subsets from databases, which include soft lenses, are designed to explore the impact of soft contact lenses on the performance: \\
\indent	1) The training and testing subset contains bona fide iris images, soft and textured contact lens attack images. In this way, the image with a soft contact lens is treated as a bona. \\
\indent	2) No images with soft lenses are used in the training phase. However, the testing data included both soft and textured contact lenses, and also bona fide images. Only the subject wearing textured contact should be detected as an attack presentation. \\
\indent	3) The training and testing subset contains only bona fide iris images and textured contact lens attack images.

The following metrics are used to measure the \gls{pad} algorithm performance:
\begin{itemize}
    \item \textbf{Correct Classification Rate (CCR)}: The ratio between the total number of correctly classified images and the number of all classified presentations. This metric follows the same defined metric in the relative \gls{sota} works \cite{multitaskcnn18, psf18, patchcnn18, desist16, opensource18}. Other works reported the Total Error rate, which is $1 - \gls{ccr}$ \cite{fusionvgg18, muipad, densepad19}.
	\item \textbf{\gls{apcer}}: The proportion of attack images incorrectly classified as bona fide samples.
	\item \textbf{\gls{bpcer}}: The proportion of bona fide images incorrectly classified as attack samples.
	\item \textbf{\gls{hter}}: corresponds to the average of \gls{apcer} and \gls{bpcer}.
\end{itemize}

To compare with the \gls{sota} iris PAD algorithms, we report the CCR, APCER, BPCER, and HTER. The threshold for each stripe is 0.5 (which is the network convergence threshold), and multiple stripes vote the final decision. To provide a more comprehensive comparison, we implement several baseline methods (where we were able to acquire sufficient reproducible information) and report the EER,  BPCER values by fixing the APCER at $0.1\%$ and $1\%$ respectively. The \gls{apcer} and \gls{bpcer} follows the standard definition presented in the ISO/IEC 30107-3 \cite{ISO301073}.

\subsection{Results}

\begin{table*}
	\begin{center}
	\scalebox{0.78}{
            \begin{tabular}{|c|c|c|c|c|c|c|c|c|c|}
            \hline
\multirow{2}{*}{Database} & \multirow{2}{*}{Metric} & \multicolumn{8}{c|}{Presentation Attack Detection Algorithm (\%)} \\ \cline{3-10} 
 & & LBP\cite{lbp14}  & WLBP \cite{wlbp10}  & DESIST \cite{desist16} & MH \cite{fusionvgg18} & VGG \cite{fusionvgg18} & MHVF \cite{fusionvgg18} & MobileNetV3  & MSA(ours)  \\ \hline
\multirow{4}{*}{NDCLD-2015 \cite{ndcld15}} & CCR & 74.42 & 76.98 & 82.48 & 85.43 & 98.92 & 98.99 & 96.54 & 99.92 \\ \cline{2-10} 
                            & ACPER & 6.15 & 50.58 & 29.81 & 21.73 & 1.54 & 1.92 & 3.24 & 0.18\\ \cline{2-10} 
                            & BPCER & 38.70 & 4.41 & 9.22  & 9.74 & 0.78 & 0.39 & 3.59  & 0.00 \\ \cline{2-10} 
                            & HTER  & 22.43 & 27.50 & 19.52 & 15.74 & 1.16 & 1.16 & 3.43 & \textbf{0.09} \\ \hline \hline
\multirow{4}{*}{NDCLD-2013 (LG4000) \cite{ndcld2013}} & CCR & 99.75 & 98.67 & 99.50 & 99.92 & 100 & 100 & 98.25  & 100 \\ \cline{2-10} 
                                     & APCER & 0.00 & 2.00 & 0.50 & 0.25 & 0.00 & 0.00 & 0.00 & 0.00  \\ \cline{2-10} 
                                     & BPCER & 0.38 & 1.00 & 0.50 & 0.00 & 0.00 & 0.00 & 2.63 & 0.00  \\ \cline{2-10} 
                                     & HTER  & 0.19 & 1.50 & 0.50 & 0.13 & \textbf{0.00} & \textbf{0.00} & 1.32 & \textbf{0.00} \\ \hline \hline
\multirow{4}{*}{NDCLD-2013 (AD100) \cite{ndcld2013}}  & CCR & 92.33 & 87.67 & 98.33  & 99.67 & 99.67 & 99.67 & 97.67 & 99.67 \\ \cline{2-10} 
                                     & APCER & 0.00 & 9.00 & 2.00  & 0.00 & 1.00 & 1.00 & 3.0 & 1.00 \\ \cline{2-10} 
                                     & BPCER & 11.50 & 14.00 & 1.50  & 0.50 & 0.00 & 0.00 & 2.0 & 0.00\\ \cline{2-10} 
                                     & HTER  & 5.75 & 11.50 & 1.75  & \textbf{0.25} & 0.50 & 0.50 & 2.5 & 0.50            \\ \hline
        \end{tabular}
    }
    \end{center}
	\caption{Iris PAD performance (\%) of our proposed \gls{msa} algorithm and existing algorithms on NDCLD-2015 and NDCLD-2013 databases. MH, VGG, and MHVF are reported from the paper \cite{fusionvgg18}, which fuse MH and VGG features.}
	\label{Tab:cld_results}
\end{table*}

\begin{table*}
	\begin{center}
	\scalebox{0.85}{
\begin{tabular}{|c|c|c|c|c|c|c|c|}
\hline
\multirow{2}{*}{LivDet-2017 Database} & \multirow{2}{*}{Metric} & \multicolumn{6}{c|}{Presentation Attack Detection Algorithm (\%)} \\ \cline{3-8} 
&  & SpoofNet \cite{spoofnet_tuning} & Best CNN \cite{crossdomain19} & VGG & LivDet-2017 Winner \cite{livedet17} & MobileNetV3 & MSA (ours) \\ \hline
\multirow{3}{*}{IIITD-WVU \cite{livedet17}} & APCER & 0.34 & 21.81 & 26.45 & 29.40 & 7.05 & 2.31 \\ \cline{2-8} 
                            & BPCER & 36.89 & 72.22 & 17.09 & 3.99 & 28.06 & 19.94  \\ \cline{2-8} 
                            & HTER  & 18.62 & 47.02 & 21.77 & 16.70 & 17.56 & \textbf{11.13} \\ \hline \hline
\multirow{3}{*}{Notre Dame \cite{livedet17}} & APCER & 18.05 & 16.56 & 10.11 & 7.78 & 30.44 & 12.28          \\ \cline{2-8} 
                            & BPCER & 0.94  & 2.44  & 2.62  & 0.28 & 4.27 & 0.17  \\ \cline{2-8} 
                            & HTER  & 9.50  & 9.50   & 6.37 & \textbf{4.03} & 17.31 & 6.23 \\ \hline
            \end{tabular}
    }
    \end{center}
	\caption{Iris PAD performance (\%) of our proposed \gls{msa} algorithm and existing algorithms on IIITD-WVU and Notre Dame database. SpoofNet \cite{spoofnet_tuning} in first column is used to fine-tuning the hyperparameters of SpoofNet, which used in the competition.} 
	\label{Tab:livdet_results}
\end{table*}

\subparagraph{Rationalizing the Micro-stripes} 

\begin{figure}[t]
	\begin{center}
		\includegraphics[width=1.0\linewidth]{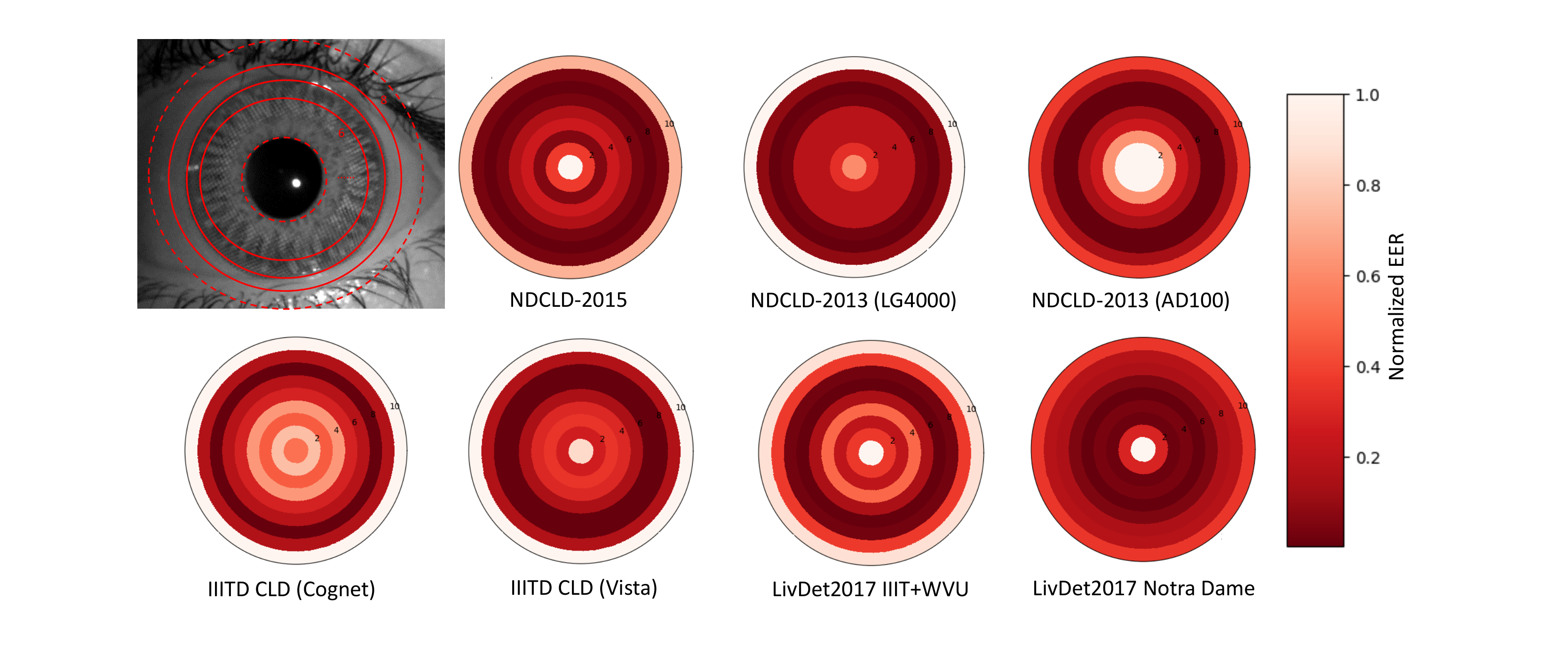}
	\end{center}
	\caption{EER heatmaps for different circular rings of segmentation. The area on the outer boundary of the iris and inner boundary of the sclera produces the lowest PAD EER (darker color) over most attacks. This points out the significance of this area for an accurate PAD. }
	\label{fig:eer_heatmap}
\end{figure}

To support our assumption that the iris/sclera boundary area is meaningful for the PAD decision, we plot the heatmap of each database based on EER values. This heat map shows the relative significance of different stripes (rings centered around the pupil) by showing the EER achieved if only one stripe is used to make a PAD decision. As shown in Fig. \ref{fig:eer_heatmap}, a normal segmentation of iris is expanded then divided into ten thinner circular rings. Then, each ring is normalized by Daugman’s rubber sheet technique \cite{daugman04} (resulting in our micro-stripe) and fed to MobileNetV3 network for training from scratch. Finally, EER values of circular rings on one database are achieved and normalized between $0-1$ to plot heatmap. In this experiment, for each database, each stripe (ring) was used separately to train (on training dataset) an individual network that was tested on the corresponding test data to measure the EER. The light color in the plot points to the highest EER values, and the darkest red points the lowest EER (and thus the most PAD-significant region). In most cases, the region from the sixth to the eighth ring has the darkest red, which yields to the lowest EER values. Besides, the ring closest to the pupil tends to be of light color, which indicates that this part of the iris does not provide much information for a PAD decision. Hence, we believe that the iris boundary comprises useful information for PAD as its region produces the most accurate PAD decisions.

\vspace{-4mm}
\paragraph{Comparison with \gls{sota}}
Tab.\ref{Tab:cld_results} and Tab.\ref{Tab:livdet_results} summarize the results per APCER, BPCER and HTER. For evaluation on NDCLD-2015 \cite{ndcld15}, NDCLD-2013 \cite{ndcld2013} , all iris images are used including the images with soft lenses (bona fide). For experiments on databases in the LivDet-2017 competition \cite{livedet17}, known and unknown test sets are combined as reported in \gls{sota} algorithms. Since the prior \gls{sota} results on IIITD CLI databases \cite{iiitd_cli_2, iiitd_cli} were only reported by CCR, we do not list in the Table. In addition, we report the EER, BPCER by fixing APCER at $0.1\%$ and $1\%$ separately (see Tab.\ref{Tab:baseline_results}). The results in both tables are reported for a micro-stripe size of $32 \times 512$ pixels, overlapping stride of $4$ pixels. The threshold for each micro-stripe is $0.5$, the final decision is voted by majority. Our MSA solution achieved superior or similar results compared with \gls{sota} algorithms. \\
To prove that our proposed method does not rely on the network structure, but rather on the micro-stripes concept, we use the same network structure and training parameters to classify the full eye image without segmentation. This method is reported as the "MobileNetV3" in Tab.\ref{Tab:cld_results}, Tab.\ref{Tab:livdet_results} and Tab.\ref{Tab:baseline_results}. Looking at Tab.\ref{Tab:livdet_results}, MHVF \cite{fusionvgg18} method, which proposed to fuse Multi-Level Haralick (MH) and VGG features for classification, obtains higher performance than the MobileNetV3. However, this might be due to the large (number of parameters) and the pre-trained nature of the VGG16 solution. Compare to VGG and MHVF methods\cite{fusionvgg18}, our proposed MSA solution outperformed all the previously reported results on the NDCLD-2015 \cite{ndcld15} and NDCLD-2013 (LG4000) \cite{ndcld2013} databases. For example, the HTER of the previously best-performing algorithm (MHVG \cite{fusionvgg18}) is decreased from $1.16\%$ to $0.09\%$ achieved by our MSA solution. Besides, we also achieve the same results with VGG and MHVF methods \cite{fusionvgg18} but slightly lower results than MH features. This can be caused by the limited training data (as we are training from scratch) in the AD100 (600 images). The CCR values reported in ContlensNet \cite{ContlensNet} are 96.91\% in NDCLD-2013 (LG4000) and 95.00\% in NDCLD-2013 (AD100), which are both lower than our 100\% and 99.67\% respectively. \\
As shown in Tab.\ref{Tab:livdet_results}, it can be observed that our MSA method achieves significant improvement on IIITD-WVU dataset. The HTER is decreased from $16.70\%$ obtained by the LivDet-2017 Iris competition winner to $11.13\%$ by our MSA solution. It should be noticed that IIITD-WVU dataset contains not only contact lens attacks but also bona fide printouts and textured lenses printouts (the same experimental setup as the reported SotA results). It can prove that our MSA solution can not only handle contact lenses attacks but also other types of presentation attacks. In \cite{crossdomain19}, they tried to combined $61$ CNN classifiers and achieved better results than the competition winner and us on the Notre Dame dataset. However, $61$ classifiers is computationally expensive (e.g. for smartphones). Also, our method achieves slightly worse HTER than the competition winner whose method is unknown. But, considering that our HTER is $6.23\%$ in comparison to  $9.50\%$ by their best CNN classifier \cite{crossdomain19}, our MSA solution is competitive.

\begin{table}[t]
	\begin{center}
	\scalebox{0.65}{
\begin{tabular}{|c|c|c|c|c|c|}
\hline
\multirow{2}{*}{Database} & \multirow{2}{*}{Metric} & \multicolumn{4}{c|}{Presentation Attack Detection Algorithm (\%)} \\ \cline{3-6} 
&  & LBP   & VGG   & MobileNetV3 & MSA(ours) \\ \hline
\multirow{3}{*}{NDCLD-2015}            & EER     & 30.64 & 9.43  & 0.37  & \textbf{0.06}      \\ \cline{2-6} 
                                       & B\_0.1 & 86.88 & 3.03  & 0.38  & \textbf{0.00}      \\ \cline{2-6} 
                                       & B\_1    & 73.75 & 2.35  & 0.38  & \textbf{0.00}      \\ \hline \hline
\multirow{3}{*}{NDCLD-2013(LG4000)}    & EER     & 24.56 & \textbf{0.00} & 1.25  & \textbf{0.00}      \\ \cline{2-6} 
                                       & B\_0.1 & 77.88 & \textbf{0.00}  & 5.5   & \textbf{0.00}      \\ \cline{2-6} 
                                       & B\_1    & 48.25 & \textbf{0.00}  & 0.38  & \textbf{0.00}      \\ \hline \hline
\multirow{3}{*}{NDCLD-2013 (AD100)}    & EER     & 17.25 & \textbf{0.00}  & 5.00  & 0.25      \\ \cline{2-6} 
                                       & B\_0.1  & 32.00 & \textbf{0.00}  & 49.00 & 0.50      \\ \cline{2-6} 
                                       & B\_1    & 30.5  & \textbf{0.00} & 48.00 & \textbf{0.00}      \\ \hline \hline
\multirow{3}{*}{IIIT CLD Cognet}       & EER     & 36.22 & 4.63  & 1.63  & \textbf{0.33}     \\ \cline{2-6} 
                                       & B\_0.1  & 99.67 & 23.70 & 12.02 & \textbf{3.58}     \\ \cline{2-6} 
                                       & B\_1    & 98.21 & 9.80  & 5.20  & \textbf{0.26}     \\ \hline \hline
\multirow{3}{*}{IIITD CLD Vista}       & EER     & 42.18 & 2.12  & 1.79  & \textbf{0.14}     \\ \cline{2-6} 
                                       & B\_0.1  & 99.80 & 9.80  & 3.67  & \textbf{0.00}      \\ \cline{2-6} 
                                       & B\_1    & 99.10 & 2.97  & 2.647 & \textbf{0.00}       \\ \hline \hline
\multirow{3}{*}{LivDet2017 IIIT+WVU}   & EER     & 23.78 & 22.98 & 18.17 & \textbf{12.38}     \\ \cline{2-6} 
                                       & B\_0.1 & 100   & 97.15 & 86.47 & \textbf{45.01}     \\ \cline{2-6} 
                                       & B\_1    & 95.44 & 90.59 & 74.50 & \textbf{29.34}     \\ \hline \hline
\multirow{3}{*}{LivDet2017 Notre Dame} & EER     & 40.84 & \textbf{5.72}  & 10.17 & 6.22      \\ \cline{2-6} 
                                       & B\_0.1 & 96.00 & \textbf{41.89} & 68.06 & 69.78     \\ \cline{2-6} 
                                       & B\_1    & 85.06 & \textbf{25.06} & 60.39 & 27.00     \\ \hline
\end{tabular}
    }
    \end{center}
	\caption{B\_0.1 and B\_1 are the BPCER value by fixing APCER at 0.1\% and 1\% respectively. Our MSA method always achieves the best or close second best on different databases and metrics.}
	\label{Tab:baseline_results}
\end{table}

\vspace{-4mm}
\paragraph{Comparison with Baselines}
In addition to comparison with \gls{sota} algorithms, we report EER and two BPCERs by fixing APCER at $0.1\%$ and $1\%$ in Tab. \ref{Tab:baseline_results}. Since most \gls{sota} algorithms, which performed experiments on these databases, evaluated performance by CCR, APCER, and BPCER values, we implement several baseline methods to provide a wider view of the performance. By observing the Tab. \ref{Tab:baseline_results}, we can find that our proposed MSA has the lowest EER, lowest BPCERs in most cases, keeping in mind the efficiency factor of MSA.

\vspace{-4mm}
\paragraph{Impact of Soft Lens}
Baker \etal \cite{degradation10} reported that soft lenses lead to the degradation of iris recognition performance. It is interesting to explore the impact of soft lenses on iris PAD performance. Hence, additional experiments are demonstrated to evaluate if our method can make a correct decision unaffected by soft lenses, i.e., do not classify soft lenses as attacks and thus increase the HTER.

Three experimental settings are described in section \ref{ssec:experiment_setup} and results are shown in Tab.\ref{Tab:soft_lense} on databases which contain soft contact lenses. Looking at Tab.\ref{Tab:soft_lense}, the results of experiment 1 using soft lenses have slightly higher HTER than experiment 3 which holds no soft contact lenses. Moreover, the results of experiment 2 where training without soft lenses and test with soft lenses achieve similar HTER values of experiment 1. For example, the HTER values by our MSA solution from Experiment 1, 2, 3 on database Cognet is $1.04\%$, $0.96\%$ and $0.66\%$ respectively by given that the percentage of the soft lens images is 33.02\% in the training set and 32.16\% in the test set. The accuracy of employing the model that never see soft contact lenses before decreases only by $0.08$ percentage points than the model who has already learned related features. This proves that our proposed MSA solution is able to classify the unknown soft lenses correctly as bona fide.  We do not compare our results with the \cite{psf18}, because the reconstructed NDCLD-2015 dataset \cite{ndcld15} by selecting some specific iris images (dot-like pattern) is manually and hard to reproduce.

\begin{table}[t]
	\begin{center}
	\scalebox{0.65}{
\begin{tabular}{|c|c|c|c|c|c|}
\hline
\multirow{2}{*}{Database} & \multirow{2}{*}{Experiment} & \multicolumn{4}{c|}{HTER (\%)} \\ \cline{3-6} 
&  & LBP   & VGG   & MobileNetV3 & MSA(ours) \\ \hline
\multirow{3}{*}{NDCLD-2015}            & 1    & 28.73  & 8.74  & 0.78  & \textbf{0.09}      \\ \cline{2-6} 
                                       & 2    & 25.34 & 6.12  & 0.37  & \textbf{0.03}      \\ \cline{2-6} 
                                       & 3    & 26.11 & 6.99  & 0.75  & \textbf{0.00}      \\ \hline \hline
\multirow{3}{*}{NDCLD-2013(LG4000)}    & 1    & 27.13 & \textbf{0.00}  & 1.32  & \textbf{0.00}      \\ \cline{2-6} 
                                       & 2    & 26.39 & \textbf{0.00}  & 0.25   & \textbf{0.00}      \\ \cline{2-6} 
                                       & 3    & 26.13 & \textbf{0.00}  & 0.75  & \textbf{0.00}      \\ \hline \hline
\multirow{3}{*}{NDCLD-2013 (AD100)}    & 1    & 21.00 & \textbf{0.00}  & 2.50  & 0.50      \\ \cline{2-6} 
                                       & 2    & 16.25 & \textbf{0.00}  & 1.01 & 1.00      \\ \cline{2-6} 
                                       & 3    & 20.50  & \textbf{0.00} & 0.73 & 1.50      \\ \hline \hline
\multirow{3}{*}{IIIT CLD Cognet}       & 1    & 42.19 & 5.02  & 3.94  & \textbf{1.04}     \\ \cline{2-6} 
                                       & 2    & 33.07 & 4.01  & 4.15 & \textbf{0.96}     \\ \cline{2-6} 
                                       & 3    & 36.63 & 4.40  & 7.69  & \textbf{0.66}     \\ \hline \hline
\multirow{3}{*}{IIITD CLD Vista}       & 1    & 43.73 & 2.40  & 0.35  & \textbf{0.00}     \\ \cline{2-6} 
                                       & 2    & 42.00 & 2.78  & 0.48  & \textbf{0.09}     \\ \cline{2-6} 
                                       & 3    & 42.04 & 3.02  & 0.60 & \textbf{0.10}     \\ \hline 
\end{tabular}
    }
    \end{center}
	\caption{Comparison of three different experiments to explore the performance influence of soft contact lenses.} The experimental setting is described in Sec. \ref{ssec:experiment_setup}. Experiment 1 is designed that both training and test stages are performed on iris subject wearing no lenses, soft lenses and textured lenses. Experiment 2 is designed that training without soft lenses but testing with soft lenses. In Experiment 3, training and test phases contain no soft lenses. In these experiments, a soft lens is considered as bona fide.
	\label{Tab:soft_lense}
\end{table}

\vspace{-4mm}
\paragraph{Impact of Overlapping Micro-Stripes}
\begin{figure}[t]
	\begin{center}
		\includegraphics[width=0.9\linewidth]{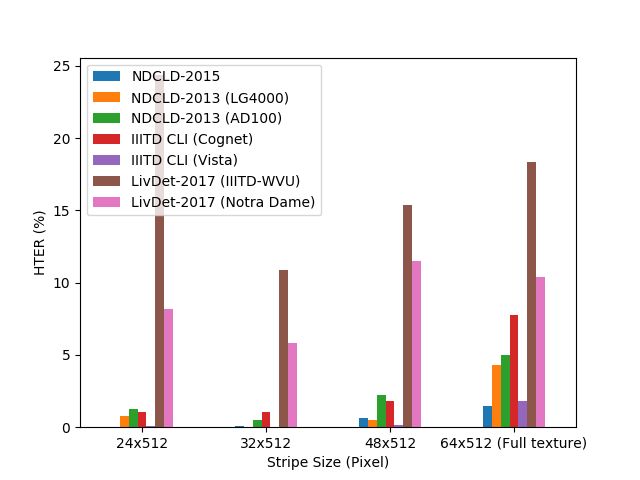}
	\end{center}
	\caption{Performance (HTER (\%)) of the different size of overlapping micro-stripes on the investigated databases. Size of $64 \times 512$ means that full normalized iris are fed to network.}
	\label{fig:stripes_hist}
\end{figure}

The purpose of this subsection is to prove the positive effect of our proposed micro-stripes approach. As a baseline, we use the full segmented area as one large stripe $64 \times 512$ pixels, processed by the same network structure and trained with the same experimental settings as the micro-stripes. We also investigate using stripes of different heights ($24$, $32$, and $48$ pixels). Figure \ref{fig:stripes_hist} illustrates the HTER achieved by these different settings. The figure \ref{fig:stripes_hist} shows that introducing the \gls{msa} approach improved the HTER on all databases, in comparison to the classifying the full segmented area. Having a micro-stripe of the height 32 pixels improved the HTER on all databases. For example, decrease the HTER from $18.36\%$ with $64 \times 512$ size to $10.88\%$ with $32 \times 512$ size on IIIT-WVU dataset and from $10.40\%$ to $5.80\%$ on Notre Dame dataset. We assume that the stripes provides the network with a chance to better generalize on the problem by providing larger amount of samples, as well as, samples with less complicated information (smaller area). One can imagine the effect of the micro-stripes in a similar manner to data augmentation, leading to lower overfitting. Since we are looking for certain image dynamic signs in the iris/sclera border area, and since the segmentation is not always optimal, these signs of attack might occur at different places. The nature of the overlapping micro-stripes is assumed to provide robustness to this localization issue. However, in some scenarios, having a very thin micro-stripes might reduce the performance, e.g. $24$ pixel stripes on the IIITD-WVU database (see Figure \ref{fig:stripes_hist}). This might be due to the fact that a very small stripe might not have sufficient information to make the detection decision. Therefore, the size $32 \times 512$ pixels of the micro-stripe should be consider suitable for higher performance on all databases.

\vspace{-5mm}
\paragraph{Impact of Majority Vote}
Once the model is trained, the next question is how to test new images. As one of our proposed MSA goals is to be robust to iris/sclera boundary localization and the localization of attack hints, we produce \gls{pad} decisions from multiple micro stripes. These multiple decisions are fused to produce a unified decision. These decisions can be joint on the decision-level by majority voting ($p_{mv}$) or on the score-level (classifier confidence) by calculating an average decision score of the micro-stripes ($\overline{p}$). We investigated both scenarios with a baseline of the full segmentation area ($64 \times 512$ pixels) passed into the micro-stripe PAD network (after resizing) ($p_{resize}$). 
Table \ref{Tab:mv} shows the result of this comparison and points out the superiority of the majority vote over other methods. This might be due to the fact that some minority stripes will be very confident that an attack is a bona fide, as it does not contain in its area any hints of an attack. All the results in this investigation considered the micro-stripe height of $32$ pixels.

\begin{table}
	\begin{center}
	    \scalebox{0.7}{
		\begin{tabular}{|c|c|c|c|}
			\hline
			\multirow{2}{*}{Database} & \multicolumn{3}{c|}{HTER (\%)} \\ \cline{2-4} 
						& $p_{mv}$  & $\overline{p}$ & $p_{resize}$   \\ \hline
			NDCLD-2015              & 0.09     &  1.02         & 1.47          \\ \hline
			NDCLD-2013 (LG4000)     & 0.00     &  1.33         & 4.31          \\ \hline
			NDCLD-2013 (AD100)      & 0.50     &  2.56         & 5.00          \\ \hline
			IIITD CLD Cognet        & 1.04     &  3.78        & 7.74          \\ \hline
			IIITD CLD Vista         & 0.18     &  0.89        & 1.83          \\ \hline
			LivDet-2017 IIITD-WVU   & 10.88    &  15.37        & 18.36          \\ \hline
			LivDet-2017 Notre Dame  & 5.80     &  7.71        & 10.40          \\ \hline
		\end{tabular}
		}
	\end{center}
	\caption{The performance of testing images with different evaluation strategies. $p_{mv}$, $\overline{p}$, $p_{resize}$ represent the accuracy with decision-level majority vote, score-level average of classification score, and the accuracy produced by the full segmentation area passed into the micro-stripe PAD network, respectively. The superiority of the majority vote is noted.}
\label{Tab:mv}
\end{table}

%-------------------------------------------------------------------------

\section{Conclusion}
Iris recognition is vulnerable to the presentation attacks, such as contact lenses and printouts. To address iris PAD issue, we propose the micro-stripe analyses (\gls{msa}) solution to detect iris presentation attack. Our solution tries to focus on the differences (between attack and bonafide) in the image dynamics around the iris/sclera border area. To do that, we propose to analyze thin micro-stripes of the normalized border area. These micro-stripes provides more samples, more consistent, and less sample dimensionality, to the training process, resulting in a better fitted model. The decision of multiple overlapping stripes are fused by a simple majority vote to build the final detection decision. 
We evaluated our method on five databases including databases in the available ones from the LivDet-2017 Iris competition. Our approach surpassed the detection performance of state-of-the-art methods in most cases, e.g., the HTER on the IIITD-WVU database was decreased to $11.13\%$, from $16.70\%$ achieved by the winner of this database in the competition. Additionally, our micro-stripe solution achieve the lowest EER and lowest BPCER values at 0.1\% and at 1\% APCER in most cases in comparison to general baselines. Furthermore, our \gls{msa} solution do not demonstrate the common issue of confusing bona fide transparent lenses with attack textured lenses.

\noindent\textbf{Acknowledgment:}
This research work has been funded by the German Federal Ministry of Education and Research and the Hessen State Ministry for Higher Education, Research and the Arts within their joint support of the National Research Center for Applied Cybersecurity ATHENE.

% reference
{\small
\bibliographystyle{ieee}
\bibliography{submission_example}
}

\end{document}